\documentclass[11pt]{article}
\usepackage{PRIMEarxiv}
\usepackage[utf8]{inputenc}
\usepackage[T1]{fontenc}
\usepackage{graphicx}
\usepackage{multirow}
\usepackage{amsmath,amssymb,amsfonts,amsthm}
\usepackage{mathrsfs}
\usepackage{xcolor}
\usepackage{textcomp}
\usepackage{booktabs}
\usepackage{tabularx}
\usepackage{placeins}
\usepackage{microtype}
\usepackage{url}
\usepackage{hyperref}
\usepackage{orcidlink}
\hypersetup{
 colorlinks=true, linkcolor=black, citecolor=blue, urlcolor=blue,
 pdfauthor={Zeynep Ozdemir, Zehra Hasgul, Irem Ulku, Omer Ozgur Tanriover, Fatma Karacaoglu, Murat Akkaya},
 pdftitle={GingivitisSeg: A Public Dataset and Comparative Benchmark for Gingivitis Segmentation in Intraoral Images}
}
\graphicspath{{./}}
\title{GingivitisSeg: A Public Dataset and Comparative Benchmark for Gingivitis Segmentation in Intraoral Images}

\author{
\textbf{Zeynep Ozdemir}$^{1,*}$\,\orcidlink{0000-0003-2985-4838},
\textbf{Zehra Hasgul}$^{2}$\,\orcidlink{0009-0002-5063-7125},
\textbf{Irem Ulku}$^{3}$\,\orcidlink{0000-0003-4998-607X}
\\[0.5em]
\textbf{Omer Ozgur Tanr\i over}$^{3}$\,\orcidlink{0000-0003-0833-3494},
\textbf{Fatma Karacao\u{g}lu}$^{2}$\,\orcidlink{0000-0003-0122-2038},
\textbf{Murat Akkaya}$^{4}$\,\orcidlink{0000-0002-3211-6572}
\\[0.8em]
\normalfont\small
$^{1}$Department of Artificial Intelligence and Data Engineering, Ankara University, \\
Ankara, T\"urkiye \\[0.3em]
$^{2}$Department of Periodontology, Faculty of Dentistry, Ankara University, \\
Ankara, T\"urkiye \\[0.3em]
$^{3}$Department of Computer Engineering, Ankara University, Ankara, T\"urkiye \\[0.3em]
$^{4}$Department of Periodontology, Faculty of Dentistry, Lokman Hekim University, \\
Ankara, T\"urkiye \\[0.7em]
\footnotesize
$^*$Corresponding author: \texttt{zynpozdemir@ankara.edu.tr} \\
\texttt{zynbzdmr@gmail.com}
}

\date{}
\begin{document}
\maketitle
\begin{abstract}

\textbf{Background:} Gingivitis segmentation from intraoral images is important for computer assisted periodontal
assessment and oral health monitoring. Although deep learning models have been increasingly
adopted in dental image analysis, existing studies have primarily focused on gingivitis
detection rather than precise segmentation of gingival inflammation at the pixel level.
Furthermore, the severe class imbalance between gingivitis foreground and background regions
makes accurate segmentation particularly challenging.

\textbf{Methods:} We conducted a comparative analysis of semantic segmentation models for gingivitis segmentation
from intraoral photographs. To facilitate reproducible benchmarking, we introduce the
GingivitisSeg dataset comprising 961 intraoral images with expert annotations of gingival
inflammation at the pixel level and predefined training, validation, and test partitions. The
dataset was divided into 672 training, 144 validation, and 145 test images. Seven
configurations representing different semantic segmentation architectures and backbones were
evaluated using the same dataset partitions, input resolution, and evaluation metrics. The
contribution of optimization components, including Cross Entropy loss with class weighting,
Dice loss, and auxiliary supervision, was examined through controlled ablation experiments.

\textbf{Results:} UPerNet with ResNet50 achieved the highest mIoU (76.49\%) and gingivitis IoU (54.91\%).
However, its performance differences from DeepLabV3+ with ResNet50 and SegFormer with MiT-B2
were not statistically significant after Holm correction. The results show that gingivitis
segmentation remains challenging because of severe foreground and background class imbalance,
low contrast gingival boundaries, illumination variations, and anatomical diversity.

\textbf{Conclusions:} The comparative benchmark identified UPerNet with ResNet50 as the configuration with the
highest numerical overlap performance. Ablation experiments indicated that combining the
evaluated optimization components improved segmentation performance within this configuration.
Additional analyses revealed that segmentation performance was associated with gingivitis
region size and decreased under stronger image blur. The proposed GingivitisSeg dataset,
benchmark splits, and source code are publicly available at
\url{https://github.com/zeynepozdemir/GingivitisSeg}.
\end{abstract}


\section{Introduction}\label{sec:introduction}

Gingival diseases, especially gingivitis, are among the most common oral health problems worldwide and may lead to more serious periodontal conditions if left untreated. Early diagnosis and regular monitoring are important for preventing disease progression and supporting oral health management. In recent years, artificial intelligence (AI) and deep learning methods have attracted increasing attention in dental imaging applications, including gingivitis detection and oral disease analysis \cite{priya2024computervision,liu2024oral,schwendicke2020artificial}.

Several previous studies have proposed deep learning based systems for gingival disease analysis using intraoral photographs. Early studies mainly focused on disease region segmentation using convolutional neural networks such as DeepLabV3+ and U-Net \cite{li2021automatic,chau2023accuracy}. More recent studies investigated mobile deployment, external validation, and systems using vision and language models to improve usability in clinical settings \cite{cheng2024developing,chau2025external,cheng2025gumagent}. In addition, recent oral image segmentation studies explored the segmentation of multiple oral conditions such as gingivitis, dental calculus, and caries using deep learning architectures \cite{liu2024oral}.

Accurate gingivitis segmentation has potential applications in computer assisted periodontal assessment. Delineation of visually identified gingival inflammation may support the quantification of affected regions and provide useful information for future studies on disease monitoring and treatment assessment \cite{chen2026novel}. Gingivitis segmentation may also contribute to the development of automated severity assessment and dental image analysis systems \cite{kuralt20203d}.

Despite these potential applications, there is still a lack of publicly available and standardized benchmark datasets specifically designed for gingivitis segmentation. Many existing studies focus on disease localization rather than pixel level gingivitis segmentation, use limited evaluation settings, or evaluate only a small number of models. Furthermore,
comparative evaluations spanning diverse modern segmentation architectures remain limited for gingivitis segmentation tasks.

To address these limitations, we present the GingivitisSeg benchmark dataset and an evaluation framework for gingivitis segmentation from intraoral images. The dataset contains expert annotations of gingival inflammation at the pixel level to support semantic segmentation research. GingivitisSeg comprises 961 intraoral images from two image sources, together with predefined training, validation, and test partitions. In addition to the dataset, we provide a comparative evaluation of several widely used semantic segmentation architectures, including FCN/U-Net \cite{ronneberger2015u}, DeepLabV3 \cite{chen2017rethinking}, DeepLabV3+ \cite{deeplabv3plus2018}, SegFormer \cite{xie2021segformer}, and UPerNet \cite{xiao2018unified}, using different backbone configurations.

The main contributions of this study are summarized as follows:

\begin{itemize}

\item We introduce
\textbf{GingivitisSeg}, a public benchmark dataset comprising 961 intraoral images from two image sources, with expert annotations of gingival inflammation at the pixel level and predefined training, validation, and test partitions.

\item We provide a standardized benchmark for gingivitis segmentation using multiple semantic segmentation architectures and backbone configurations.

\item We analyze the strengths and limitations of different segmentation approaches through quantitative and qualitative evaluations together with paired statistical comparisons and an analysis of training convergence.

\item We investigate the effectiveness of
established optimization components, including class weighting, Dice loss, and auxiliary supervision, through controlled ablation experiments.

\item We further examine the relationship between gingivitis region size and segmentation performance, the effects of preserving image aspect ratio during resizing, and model sensitivity to Gaussian and motion blur.

\end{itemize}


\begin{figure}[htbp]
\centering
\fbox{\includegraphics[width=0.985\linewidth]{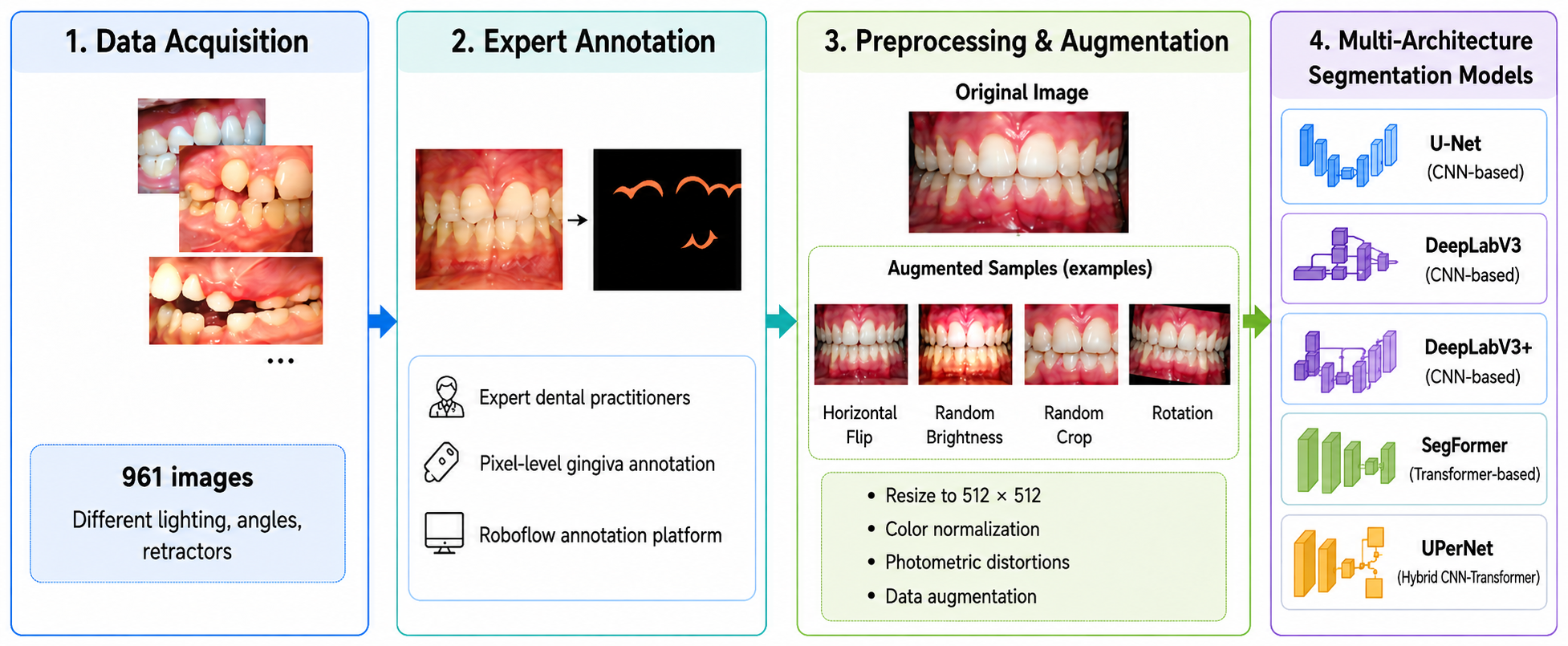}}
\caption{
Overview of the proposed experimental workflow for
gingivitis segmentation.
}
\label{fig:gingivitis_dataset}
\end{figure}


\section{Related Work}\label{sec:relatedworks}

\subsection{Dental Image Analysis}

Artificial intelligence and deep learning methods have recently attracted increasing attention in dental image analysis and computer assisted oral healthcare applications. Previous studies investigated various tasks including dental caries detection, gingivitis analysis, oral disease screening, and segmentation of intraoral structures using photographic and endoscopic images.

Several studies focused specifically on gingivitis and gingival disease detection from intraoral photographs. Li et al.~\cite{li2021automatic} proposed a deep learning framework for site specific gingival disease detection using semantic segmentation techniques. Chau et al.~\cite{chau2023accuracy} further investigated gingivitis detection from photographs using AI and evaluated the diagnostic performance of deep learning models in clinical settings. However, these studies mainly focused on  disease detection, localization, and diagnostic performance rather than systematic comparisons of multiple segmentation architectures under a common benchmark protocol, while standardized benchmark evaluation for pixel level gingivitis segmentation remained limited.

More recent works extended these systems toward practical applications and deployment. Cheng et al.~\cite{cheng2024developing} explored lightweight gingivitis detection systems suitable for mobile devices. Chau et al.~\cite{chau2025external} evaluated the external validity of gingivitis detection systems assisted by AI in older adult populations using smartphone images collected in clinical environments. In addition, Cheng et al.~\cite{cheng2025gumagent} proposed a framework named GumAgent that uses a vision and language model (VLM) to support image validation and user interaction during oral health monitoring.

Besides studies focusing on gingivitis, oral image analysis has also been investigated for multiple oral conditions. Liu et al.~\cite{liu2024oral} proposed a deep learning framework for the segmentation of gingivitis, dental calculus, and dental caries using oral endoscopic images  in a setting that addresses multiple oral diseases.

Duy et al.~\cite{duy2024dental} introduced a publicly available dataset comprising 1,096 intraoral images for gingivitis image captioning research. The dataset provides gingivitis labels based on the Modified Gingival Index (MGI) and descriptive captions. However, it was primarily developed for gingivitis assessment and image captioning rather than for comparative evaluation of segmentation models using pixel level annotations.

Although these studies demonstrate the potential of deep learning models, much of the existing literature focuses on disease detection, deployment, image captioning, or clinical validation rather than systematic evaluation of gingivitis segmentation models using a common benchmark.

\begin{table}[!htbp]
\centering
\caption{Comparison of existing studies on AI applications for gingival and oral disease analysis.}
\label{tab:dataset_comparison}

\resizebox{\textwidth}{!}{
\small
\renewcommand{\arraystretch}{1.15}

\begin{tabular}{p{3.2cm} p{2cm} p{2.8cm} p{1.5cm} p{1.8cm} p{2.8cm} p{2cm}}

\toprule

\textbf{Study} &
\textbf{Image Type} &
\textbf{Task} &
\textbf{Size} &
\textbf{Annotation} &
\textbf{Focus} &
\textbf{Evaluation} \\

\midrule

Li et al. (2021) \cite{li2021automatic} &
Intraoral Photo &
Gingivitis Seg. &
337 + 110 &
Pixel-level &
Disease Localization &
Single Model \\

Chau et al. (2023) \cite{chau2023accuracy} &
Frontal Photo &
Gingivitis Detection &
567 &
Pixel-level &
Clinical Diagnosis &
Single Model \\

Cheng et al. (2024) \cite{cheng2024developing} &
Intraoral Photo &
Mobile Gingivitis Detection &
$\sim$415 &
Pixel-level &
Mobile Deployment &
Deployment \\

Chau et al. (2025) \cite{chau2025external} &
Smartphone Photo &
Clinical Detection &
44 Patients &
Clinical Assessment &
External Validation &
Clinical Validation \\

Cheng et al. (2025) \cite{cheng2025gumagent} &
Mixed Oral Images &
Segmentation + Filtering &
332 + 83 &
Pixel-level &
VLM Assisted Workflow &
Single Model \\

Liu et al. (2024) \cite{liu2024oral} &
Endoscopic Oral Image &
Multi disease Seg. &
3365 &
Pixel-level &
Oral Disease Analysis &
U-Net Comparison \\

Duy et al. (2024) \cite{duy2024dental} &
Intraoral Photo &
Gingivitis Image Captioning &
1096 &
MGI Labels + Captions &
Gingivitis Description &
Dataset Resource \\

\midrule

\textbf{GingivitisSeg (Ours)} &
\textbf{Intraoral Photo} &
\textbf{Gingivitis Segmentation} &
\textbf{ 961} &
\textbf{Pixel-level} &
\textbf{Benchmark Evaluation} &
\textbf{Seven Configurations} \\

\bottomrule

\end{tabular}
}

\end{table}

\subsection{Medical Image Segmentation}

Semantic segmentation is an important task in medical image analysis and computer vision. Convolutional neural networks with encoder and decoder structures have been widely adopted for biomedical image segmentation because they can capture local details and contextual information.

U-Net~\cite{ronneberger2015u} is one of the earliest and most influential segmentation architectures and has been extensively used in medical imaging applications because of its encoder and decoder structure with skip connections.

DeepLabV3~\cite{chen2017rethinking} uses atrous convolution and atrous spatial pyramid pooling
(ASPP) to capture contextual information at multiple scales, while
DeepLabV3+~\cite{deeplabv3plus2018} extends this approach with an encoder and decoder design
that supports the refinement of segmentation boundaries.

More recent developments in semantic segmentation include architectures based on transformers
and modern convolutional networks. SegFormer~\cite{xie2021segformer} combines a hierarchical
transformer backbone with an efficient segmentation decoder. UPerNet~\cite{xiao2018unified}
integrates features from multiple spatial scales for dense prediction tasks and can be used
with different backbone architectures. ConvNeXt~\cite{liu2022convnet} modernized convolutional
network design by incorporating architectural principles inspired by vision transformers.

These architectures have shown promising results in general semantic segmentation benchmarks
and medical image analysis tasks. However, their effectiveness for gingivitis segmentation has
not yet been comprehensively investigated through comparisons using consistent data partitions
and evaluation criteria.

\subsection{Gingivitis Segmentation Studies}
\label{sec:gingiva_segmentation_studies}

Automatic gingivitis segmentation remains a relatively underexplored topic compared with other medical image segmentation applications. Existing studies have primarily focused on disease localization, gingivitis grading, or clinical diagnosis rather than detailed segmentation of gingivitis regions at the pixel level.

Early studies mainly employed convolutional neural network architectures for gingival disease analysis. Li et al.~\cite{li2021automatic} proposed a semantic segmentation framework based on DeepLabV3+ for gingival disease detection at multiple levels using intraoral photographs. Similarly, Chau et al.~\cite{chau2023accuracy} investigated gingivitis detection from photographs using deep learning methods and evaluated diagnostic performance metrics such as sensitivity and specificity.

Recent studies have extended gingival image analysis toward deployment and clinical applications. Cheng et al.~\cite{cheng2024developing} proposed a lightweight gingivitis detection framework for mobile devices, while Cheng et al.~\cite{cheng2025gumagent} introduced a workflow using a vision and language model for image filtering and oral health monitoring. In addition, Chau et al.~\cite{chau2025external} performed external clinical validation of gingivitis detection systems using smartphone images collected in clinical environments.

Public image resources focusing on gingivitis have also become available. In particular, Duy et al.~\cite{duy2024dental} released an intraoral image dataset containing gingivitis labels assigned by experts according to the Modified Gingival Index and descriptive captions. Although this resource increases the availability of gingivitis images, its annotation objectives differ from those of semantic segmentation because it does not provide masks identifying inflammation at the pixel level.

Despite these advances, publicly available benchmark datasets specifically designed for gingivitis segmentation remain limited. Furthermore,

comparative evaluations involving different segmentation architectures under consistent data partitions and evaluation criteria remain limited.

Therefore, the GingivitisSeg benchmark provides annotations of gingivitis regions at the pixel level together with predefined dataset partitions to support reproducible evaluation and future research on gingivitis segmentation.

\section{Proposed GingivitisSeg Dataset}\label{sec:dataset}

\begin{figure}[htbp]
\centering
\fbox{\includegraphics[width=0.985\linewidth]{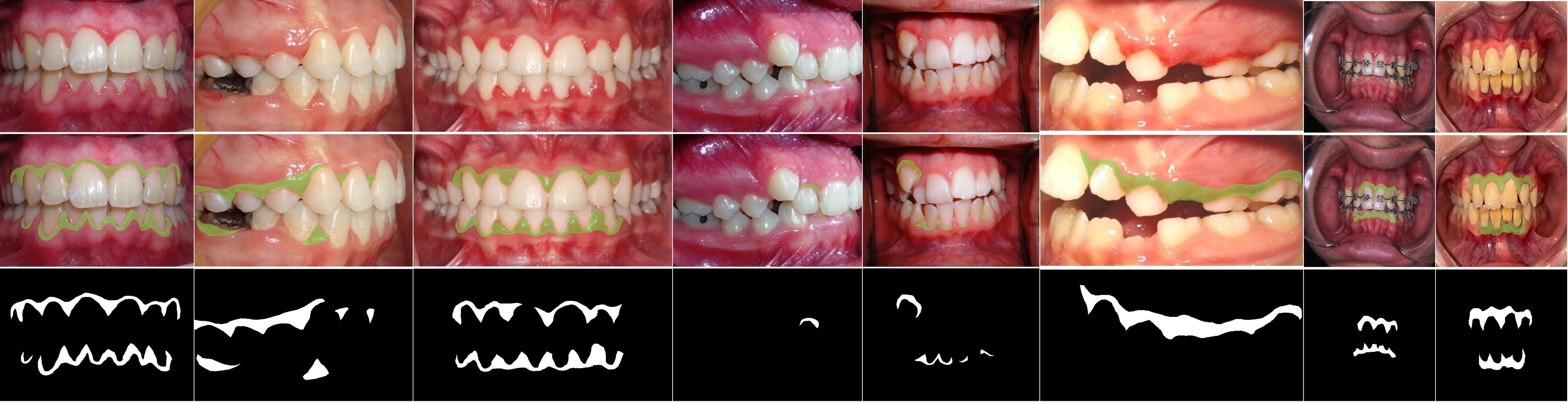}}
\caption{Representative samples from the proposed GingivitisSeg dataset.
First row: original intraoral images. Second row: expert annotations
overlaid on the original images. Third row: corresponding binary
gingival inflammation masks.}
\label{fig:dataset_samples}
\end{figure}

The proposed GingivitisSeg dataset was created to support the
development and evaluation of automatic gingivitis segmentation
methods from intraoral photographs. The dataset focuses on images
containing gingivitis and aims to provide a benchmark resource
for dental image analysis research. Representative samples
illustrating the diversity of the dataset are shown in
Fig.~\ref{fig:dataset_samples}.

Unlike the multi-disease setting of Liu et al.~\cite{liu2024oral},
the present benchmark is specifically designed for binary
pixel-level delineation of gingivitis and provides predefined
source-stratified training, validation, and test partitions
for controlled model comparison. The aim is therefore not to
claim broader clinical coverage based on dataset size, but
to provide a focused and reproducible benchmark for
gingivitis segmentation.

\subsection{Data Acquisition}\label{sec:data_acquisition}

The developed GingivitisSeg benchmark dataset combines images
from two sources. The original clinical subset contains 432
intraoral photographs collected at the Ankara University Faculty
of Dentistry during routine dental examinations. To increase
image-source diversity, this collection was supplemented with
images from the publicly available gingivitis image dataset
described by Duy et al.~\cite{duy2024dental}. The source dataset
provides gingivitis severity-level annotations but does not
contain pixel-level gingivitis segmentation masks. Following
quality control, 529 images from this public source were
retained and newly annotated for pixel-level gingivitis
segmentation by the dental experts involved in the present
study. The resulting expanded benchmark therefore comprises
a total of 961 intraoral images.

The original clinical subset comprises 432 intraoral
photographs obtained from 432 different patients presenting
with periodontal complaints to the Department of
Periodontology, Ankara University Faculty of Dentistry,
between February 2022 and February 2023. Patients with
clinically evident gingival inflammation were included
regardless of age or sex. Photographs were routinely
acquired using smartphone cameras before periodontal
treatment under clinical conditions. All photographs
obtained during the study period from patients who
consented to clinical photography were included without
further selection; refusal to provide consent for
photography was the only exclusion criterion.
Demographic information, including age and sex,
was not recorded.

These acquisition and consent procedures refer specifically
to the 432-image clinical subset collected at Ankara
University. The additional 529 images were obtained from
an existing public dataset and were therefore used as
secondary data; details of their original acquisition
procedure are reported in the corresponding source
publication~\cite{duy2024dental}.

The dataset includes images with considerable variability
in terms of imaging conditions and anatomical presentation.
While some images focus on localized gingivitis, others
capture larger parts of the oral cavity including teeth,
lips, and surrounding soft tissues. In several cases,
cheek retractors were used to expose the inflamed gingival
regions more clearly. Consequently, the dataset contains
images obtained from different viewpoints and under varying
illumination conditions.

The expanded multi-source dataset exhibits a substantially
broader range of native image resolutions. Image widths
range from 416 to 7286 pixels and image heights range
from 218 to 4863 pixels. Across all 961 images, the
mean width and height are approximately 4346 and
2894 pixels, respectively. This variability reflects
the heterogeneous acquisition characteristics of the
two image sources.

\subsection{Annotation Process}\label{sec:annotation_process}

Pixel-level gingivitis annotations were created using
the Roboflow annotation platform by two dental experts from the Department of Periodontology.
One annotator had more than 10 years of clinical
experience, while the other was a postgraduate resident.
Gingival inflammation was identified based on visible
signs in the intraoral photographs, including redness,
edema, and changes in gingival contour. Inflamed regions
were delineated as a binary segmentation class without
grading inflammation severity. 
Clinical periodontal indices, including the
Gingival Index (GI), Modified Gingival Index
(MGI), and bleeding on probing (BOP),
were not used to establish or validate
the pixel-level segmentation masks.

Before annotation, the two dental experts jointly reviewed
representative photographs to calibrate the visual criteria
for identifying gingival inflammation. Following calibration,
the photographs were divided between the annotators, and
each expert annotated a separate subset.

For the additional 529 public-source images, the original
dataset contained gingivitis severity-level labels but no
pixel-level segmentation masks. Therefore, pixel-level
gingivitis masks were newly created for the present study
by the same two dental experts. The same segmentation
target definition was used for both image sources to
maintain consistency across the expanded benchmark.

During the annotation process, particular attention was
paid to accurately identifying the boundaries of gingival
inflammation areas while distinguishing them from adjacent
oral structures such as teeth, lips, and oral mucosa.
Fine anatomical details and complex gingival boundaries
were preserved to ensure anatomically meaningful
segmentation masks.

Following the initial annotation process, all segmentation
masks were jointly reviewed by both annotators. Any
disagreements regarding gingival boundaries were resolved
through discussion until a consensus annotation was reached,
thereby improving annotation consistency and the overall
reliability of the ground-truth masks.

Although all segmentation masks underwent joint expert
review and consensus, the same images were not
independently annotated by both experts before this
process. Therefore, quantitative interobserver agreement
could not be calculated.

To assess intraobserver reliability, a subset of
30 photographs was re-annotated by the same periodontist
one month after the initial annotation. The
intraclass correlation coefficient (ICC) was 0.80,
indicating good intraobserver reliability.

The final annotations were exported from the annotation
platform in JSON format and subsequently converted into
segmentation masks compatible with the training pipelines
used in this study.

\subsection{Dataset Description}\label{sec:dataset_statistics}

The expanded GingivitisSeg dataset contains a total of
961 intraoral images, comprising 432 images from the
original clinical collection and 529 images from the
additional public source, together with their
corresponding expert-annotated pixel-level
segmentation masks.

All images contain expert-annotated gingival inflammation
regions, ensuring that the dataset specifically targets
clinically relevant gingivitis segmentation tasks.
A summary of the main dataset characteristics is
presented in Table~\ref{tab:dataset_statistics}.

Quantitative analysis of the 961 segmentation masks
confirmed a pronounced foreground--background class
imbalance. The mean image-wise gingivitis foreground
proportion was 3.51\%, whereas the corresponding mean
background proportion was 96.49\%. Across individual
images, gingivitis foreground coverage ranged from
approximately 0.12\% to 19.83\%.

This class imbalance highlights the challenging nature
of the gingivitis segmentation problem.

The dataset exhibits variability in the visual appearance and extent of gingival inflammation,
imaging angle, illumination conditions, and anatomical structures. Formal clinical severity
grading was not performed for the original clinical subset. The collected images include both
frontal and lateral intraoral views, as well as cases containing orthodontic appliances,
missing teeth, dental restorations, and varying visual manifestations of gingival inflammation.
This diversity provides a realistic and challenging benchmark for evaluating the robustness and
generalization capability of state-of-the-art semantic segmentation algorithms.

During dataset preparation, quality-control procedures
were performed. Images with missing or unusable
annotations, severe image-quality issues, unreadable
files, or formatting problems were excluded from
the final dataset. In addition, images captured in
horizontal orientations were rotated to maintain
a consistent orientation throughout the dataset.

A total of 533 candidate images were examined from
the additional public source. Four images without
usable segmentation annotations were excluded during
dataset preparation, leaving 529 public-source images
in the final benchmark. Before construction of the
final splits, duplicate screening was also performed.
No exact duplicate images or cross-source perceptual-hash
duplicate candidates were identified.

All images included in the benchmark contain annotated
gingivitis regions. Accordingly, GingivitisSeg is designed
to evaluate pixel-level delineation of gingivitis within
positive cases rather than image-level discrimination
between healthy and gingivitis-positive subjects.
This characteristic should be considered when interpreting
the scope and generalizability of the benchmark.

The dataset, together with its pixel-level annotations
and predefined benchmark splits, is made publicly
available to researchers for comparable automatic
gingivitis segmentation studies.

\subsection{Train, Validation and Test Split}\label{sec:data_split}

To enable fair and reproducible benchmarking of segmentation
models, the dataset was divided into training, validation,
and test subsets following the preprocessing stage.
The distribution of the dataset across these subsets
is summarized in Table~\ref{tab:dataset_statistics}.
The training subset is used for model optimization,
while the validation subset is used during training
for hyperparameter tuning. The test subset is reserved
exclusively for the final evaluation in order to ensure
unbiased performance assessment.

The expanded dataset was partitioned using deterministic
sample-level splitting performed separately within each
image source, thereby maintaining representation of
both sources across the training, validation, and test
subsets. The resulting benchmark contains 672 training
images, 144 validation images, and 145 test images,
corresponding to approximately 70\%, 15\%, and 15\%
of the complete dataset, respectively.

Patient-level grouping could not be applied to the complete
expanded dataset because reliable patient identifiers
were not available for the public-source images.
To reduce the risk of image-level data leakage,
duplicate screening was performed before splitting,
and no exact or perceptual cross-source duplicate
candidates were detected. The expanded benchmark should
therefore be interpreted as a source-stratified
sample-level evaluation rather than a fully
patient-disjoint evaluation.

The predefined benchmark split is released together
with the dataset to ensure fair, transparent,
and reproducible comparison across studies.


\begin{table}[htbp]
\centering
\caption{Summary statistics of the proposed GingivitisSeg dataset.}
\label{tab:dataset_statistics}
\small
\renewcommand{\arraystretch}{1.15}

\begin{tabular}{p{5.0cm}p{6.2cm}}
\toprule
\textbf{Property} & \textbf{Value}\\
\midrule

Number of images &
 961\\

Source composition &
432 clinical + 529 public-source images\\

Image type &
Intraoral photographs\\

Classes &
Background, Gingivitis Area\\

Annotation type &
Pixel-level masks\\

Native image width &
416--7286 pixels\\

Native image height &
218--4863 pixels\\

Mean image resolution &

approximately $4346\times2894$\\

Mean gingival inflammation area &

3.51\%\\

Mean background area &

96.49\%\\

Train split &

672 (69.9\%)\\

Validation split &

144 (15.0\%)\\

Test split &

145 (15.1\%)\\

\bottomrule
\end{tabular}
\end{table}

\subsection{Dataset Challenges}\label{sec:dataset_challenges}

Although the dataset focuses on gingivitis, gingivitis
segmentation remains challenging due to several factors
present in real clinical images. Variations in illumination
and reflections caused by saliva or clinical lighting may
introduce bright highlights and uneven brightness across
the image.

Furthermore, inflamed gingival regions often exhibit
color similarities with surrounding oral structures such
as lips, oral mucosa, and teeth, which can make boundary
detection difficult for segmentation algorithms.
The severity of gingivitis may also vary significantly
between patients, resulting in different color intensities
and textures.

Occlusions caused by lips, cheeks, or dental instruments
may partially obscure gingivitis areas. In addition,
variations in tooth alignment, dental restorations,
and surrounding anatomical structures introduce complex
visual patterns that segmentation models must learn
to distinguish.

The multi-source nature of the expanded benchmark
additionally introduces variation in native image
resolution and acquisition characteristics. Moreover,
gingivitis regions occupy only a small fraction of most
images, with a mean foreground coverage of 3.51\%,
making small and spatially localized inflammatory
regions particularly challenging to segment.

These factors make GingivitisSeg a challenging and
clinically relevant benchmark for evaluating the
robustness and generalization capability of
state-of-the-art deep learning-based gingivitis
segmentation methods.

\section{Methodology}\label{sec:methodology}

\begin{figure}[htbp]
\centering
\fbox{\includegraphics[width=0.985\linewidth]{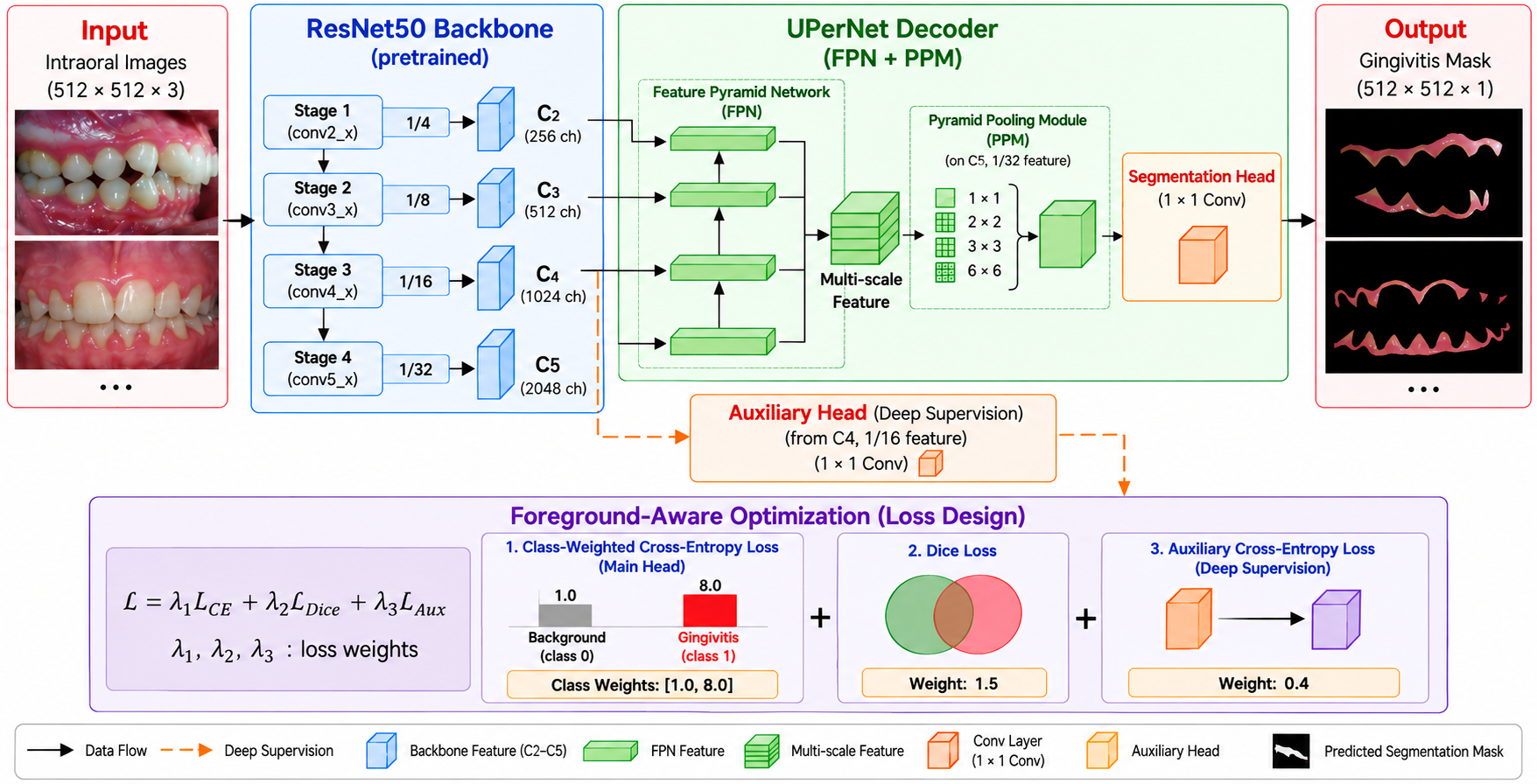}}
\caption{
Overview of the UPerNet configuration and the optimization
components evaluated in this study. The framework combines
feature aggregation at multiple scales with Cross Entropy
loss using class weighting, Dice loss, and auxiliary
supervision for gingivitis segmentation.
}
\label{fig:framework}
\end{figure}

Figure~\ref{fig:framework} illustrates the UPerNet configuration
used to examine the contribution of the optimization components.
The framework consists of four stages:
(i) image preprocessing,
(ii) feature extraction using a pretrained backbone,
(iii) semantic segmentation through the UPerNet decoder, which
integrates features at multiple spatial scales, and
(iv) optimization using the selected loss components
and auxiliary supervision.

For the comparative benchmark, all architectures were evaluated
using the same dataset partitions, network input size,
test set, and evaluation metrics. The same data augmentation
strategy was applied across the seven configurations.
However, optimization and inference settings were adapted
to the requirements of individual architectures where necessary.

\subsection{Segmentation Models}

Seven configurations representing different segmentation
architectures and backbones were evaluated:
FCN with a U-Net architecture \cite{ronneberger2015u},
DeepLabV3 with ResNet50 \cite{chen2017rethinking,he2016deep},
DeepLabV3+ with ResNet50
\cite{deeplabv3plus2018,he2016deep},
SegFormer with MiT-B0 and MiT-B2 \cite{xie2021segformer},
and UPerNet with ResNet50 and ConvNeXt-Tiny
\cite{xiao2018unified,he2016deep,liu2022convnet}.

These configurations were selected to represent
conventional convolutional networks,
segmentation architectures based on transformers,
and approaches that integrate features
from multiple spatial scales.

Among the evaluated architectures, U-Net uses
an encoder and decoder structure with skip connections
to preserve spatial information. This property is
particularly relevant to biomedical image segmentation,
where fine anatomical structures must be delineated.
In gingivitis segmentation, skip connections may help
preserve local boundary details between inflamed
gingival regions and adjacent oral structures,
including teeth, lips, and oral mucosa.

DeepLabV3 uses atrous convolution and Atrous Spatial
Pyramid Pooling (ASPP) to capture contextual information
at different receptive field sizes.
This design provides a convolutional baseline
for evaluating gingivitis regions that vary
in size and spatial extent.

DeepLabV3+ extends DeepLabV3 by combining contextual
features obtained through ASPP with an encoder and
decoder structure that incorporates features from
earlier network layers. This design supports
the refinement of segmentation boundaries.
It may be useful for gingivitis segmentation because
the size and shape of inflamed regions vary with
imaging distance, patient anatomy, and
the visible extent of inflammation.

SegFormer is a lightweight architecture based on
transformers. It combines hierarchical transformer

backbones
with an efficient decoder to model contextual
information while maintaining computational efficiency.
For gingivitis segmentation, its ability to capture
relationships between distant image regions may help
distinguish inflamed gingiva from visually similar
surrounding tissues, particularly under variations
in illumination and color.

Finally, UPerNet integrates features at multiple
spatial scales using a feature pyramid.
To investigate the influence of

backbone representations,
both ResNet50 \cite{he2016deep} and ConvNeXt-Tiny
\cite{liu2022convnet}

backbones
were evaluated.

ResNet50 provides a conventional convolutional
network baseline, while ConvNeXt-Tiny represents
a modern convolutional architecture incorporating
design principles inspired by vision transformers.
By combining local and contextual information
at different scales, UPerNet may support
the delineation of gingival boundaries in
complex intraoral photographs.

Together, the selected configurations allow
different segmentation approaches to be compared
using a common evaluation protocol.
The inclusion of two SegFormer variants and
two UPerNet variants also enables comparisons
between different backbones within
the corresponding segmentation architectures.

\subsection{Training Strategy}

\subsubsection{Image Preprocessing}

All configurations used an input size of
$512\times512$ pixels at the network interface.
Geometric transformations were applied consistently
to the RGB images and their corresponding masks
to preserve spatial alignment.
Bilinear interpolation was used when resizing
RGB images, whereas nearest neighbor interpolation
was used for segmentation masks to preserve
their discrete class labels.

Images were normalized using the ImageNet
mean and standard deviation before training.

For the UPerNet configuration with a ResNet50 backbone,
which served as the reference for the ablation and
sensitivity experiments, the main preprocessing
procedure resized images directly to
$512\times512$ pixels without preserving
their original aspect ratio.
The effect of this procedure was examined
in a separate experiment comparing direct
resizing with resizing and padding that
preserved the original aspect ratio.

\subsubsection{Data Augmentation}

To improve robustness to variations encountered
in clinical images, data augmentation was used
during training.

The same augmentation strategy was applied across all seven evaluated configurations. The
augmentation procedure consisted of random resizing, random horizontal flipping, and
photometric distortion. Images were resized using scaling factors ranging from 0.75 to 1.50,
followed by resizing to the required network input size. Horizontal flipping was applied with a
probability of 0.5 to reduce sensitivity to image orientation. Photometric distortion was also
applied to simulate variations in brightness, contrast, saturation, and color distribution
commonly encountered in intraoral photographs.

\subsubsection{Foreground Optimization}

One of the main challenges of GingivitisSeg is the pronounced imbalance between the foreground
and background classes. Gingivitis regions occupy only a small proportion of image pixels. To
address this imbalance, the evaluated optimization configurations use a combination of Cross
Entropy loss with class weighting, Dice loss, and auxiliary supervision where supported by the
architecture.

The optimization objective can be expressed as

\begin{equation}
L =
\lambda_{1}L_{CE}
+
\lambda_{2}L_{Dice}
+
\lambda_{3}L_{Aux},
\end{equation}

where $L_{CE}$ denotes the Cross Entropy loss
with class weighting calculated from the main
segmentation output, $L_{Dice}$ represents
Dice loss, and $L_{Aux}$ denotes the auxiliary
Cross Entropy loss calculated from an
intermediate segmentation output when available.

The coefficients $\lambda_{1}$, $\lambda_{2}$, and $\lambda_{3}$ determine the contribution of
each component. For architectures without auxiliary supervision, the corresponding auxiliary
term is omitted. For configurations using class weighting, the weights assigned to the
background and gingivitis classes were $[1.0,\,8.0]$, respectively. The greater weight assigned
to gingivitis increases the contribution of this less frequent class during optimization. Dice
loss was included with a weight of 1.5 to encourage overlap between predicted gingivitis
regions and the reference masks.

The contribution of these established
optimization components was examined
through a controlled ablation study
using UPerNet with a ResNet50 backbone.
The complete configuration was compared
with alternatives that excluded class weighting,
Dice loss, or auxiliary supervision separately.
A configuration using only standard
Cross Entropy loss was also evaluated
as a baseline.

\subsubsection{Auxiliary Supervision}

For architectures supporting auxiliary supervision, an additional segmentation head was
connected to intermediate feature representations during training. The auxiliary branch was
optimized using Cross Entropy loss with class weighting and a loss coefficient of 0.4. This
branch provided an additional training signal to intermediate features and supported gradient
propagation during optimization.

Architectures without an auxiliary
segmentation branch were trained without
the corresponding loss term.
Therefore, auxiliary supervision was
not used identically across all
benchmark configurations. Its inclusion
depended on the design of each architecture.


\section{Experimental Setup}
\label{sec:experimental_setup}

\subsection{Experimental Settings}

To ensure consistent evaluation, all segmentation configurations used the same training,
validation, and test partitions, network input size, and evaluation metrics. Training settings
were adapted to individual architectures where necessary. Pretrained weights were used for the
corresponding backbones when available. Model optimization was performed using the AdamW
optimizer with initial learning rates ranging from $6\times10^{-5}$ to $1\times10^{-4}$.

Polynomial learning rate scheduling was used throughout the benchmark, with scheduler
configurations adapted to individual architectures and linear warmup where applicable. Gradient
clipping was applied to support optimization stability. Training was performed using a batch
size of two.

For the final comparative experiments, all seven configurations were trained for 5,000
iterations. Validation performance was monitored periodically during training, and the
checkpoint with the highest validation mIoU was selected for evaluation on the test set. Each
configuration was trained in three independent runs. Aggregate performance metrics and results
for individual classes are reported as mean $\pm$ standard deviation across these runs.

Inference was performed on complete images for DeepLabV3, DeepLabV3+, SegFormer, and UPerNet.
The FCN/U-Net configuration used a sliding window with a crop size of $512\times512$ pixels.
All predictions were evaluated against the reference masks from the same predefined test set
using identical evaluation metrics. All experiments were conducted on a workstation equipped
with an NVIDIA GeForce GTX 1080 Ti GPU.

\begin{figure}[htbp]
\centering
\includegraphics[width=0.92\textwidth]
{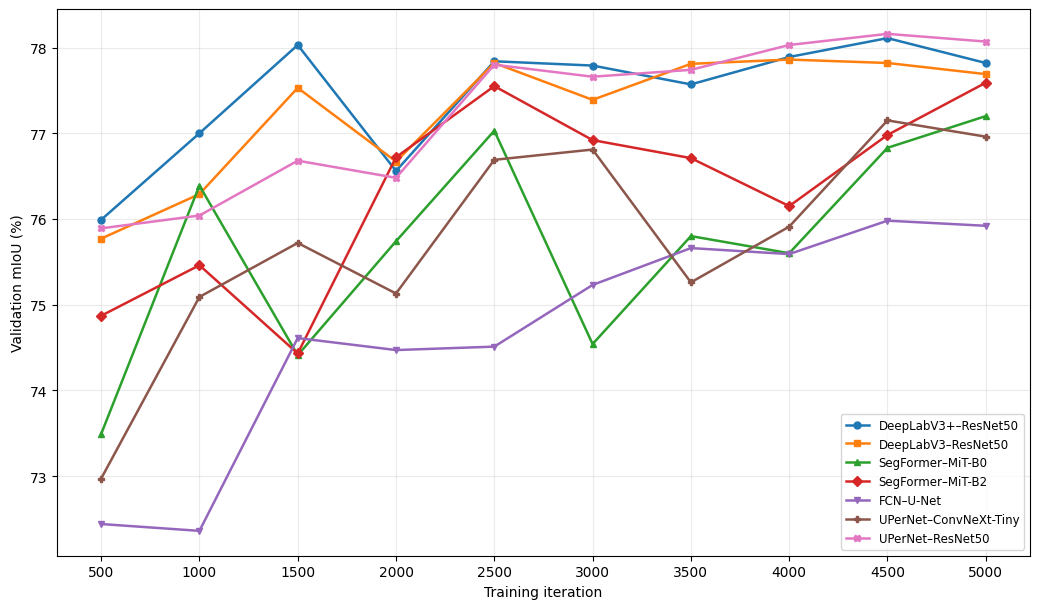}
\caption{
Validation mIoU curves for the seven segmentation
configurations over 5,000 training iterations.
Performance was recorded every 500 iterations.
The curves illustrate changes in validation
performance during training, including the
smaller fluctuations observed during later iterations.
}
\label{fig:convergence}
\end{figure}

To examine training convergence, validation
mIoU was recorded every 500 iterations during
the 5,000 training iterations for all seven
configurations.
As shown in Fig.~\ref{fig:convergence},
validation performance generally improved
rapidly during the initial training stages,
followed by smaller changes during later
iterations.
The highest recorded validation mIoU occurred
between iterations 4,000 and 5,000 for all
configurations. Although both SegFormer variants
achieved their numerical maxima at the final
evaluation point, comparable validation results
had already been observed at earlier iterations.
These observations indicate that validation
performance had become relatively stable
toward the end of training for the evaluated
configurations. However, they do not establish
that every architecture had fully converged.

\subsection{Evaluation Metrics}

To evaluate segmentation performance, commonly
used semantic segmentation metrics were employed,
including mean Intersection over Union (mIoU),
mean class accuracy (mAcc), mean Dice (mDice),
and overall pixel accuracy (aAcc).

IoU and Dice measure the agreement between predicted segmentation regions and their
corresponding reference masks. The mean metrics, mIoU and mDice, were calculated by averaging
the respective scores across the background and gingivitis classes. Mean class accuracy (mAcc)
summarizes the classification accuracy of the individual classes, whereas overall pixel
accuracy (aAcc) represents the proportion of correctly classified pixels. Because gingivitis
occupies a relatively small proportion of the images, overall pixel accuracy may be strongly
influenced by the background class. We therefore also report IoU, Dice, and class accuracy
separately for gingivitis to provide a clearer assessment of segmentation performance on the
less frequent class.

Statistical comparisons between configurations were performed using segmentation scores
calculated separately for each of the 145 test images. For these comparisons, predictions were
obtained using checkpoints selected according to validation mIoU. UPerNet with a ResNet50
backbone was chosen as the reference configuration because it achieved the highest numerical
mIoU and mDice in the benchmark. Two-sided paired Wilcoxon signed-rank tests were applied to
mIoU, mDice, gingivitis IoU, and gingivitis Dice calculated for individual test images. For
each metric, the resulting p-values were adjusted for multiple comparisons using the Holm
procedure. An adjusted $p<0.05$ was considered statistically significant. These tests assess
differences between paired predictions on the evaluated test images. The results were
interpreted together with the mean and standard deviation obtained from the three independent
training runs.

\section{Results}\label{sec:results}

Table~\ref{tab:gingiva_benchmark} presents the performance of the
evaluated segmentation configurations on the GingivitisSeg dataset.
Differences were observed between models, particularly in metrics
measuring the overlap between predicted and reference regions.


\begin{table}[htbp]
\centering
\color{black}
\caption{
Benchmark comparison of semantic segmentation models on the
GingivitisSeg dataset. Results are reported as mean $\pm$
standard deviation. All metrics are reported as percentages.
}
\footnotesize
\label{tab:gingiva_benchmark}

\begin{tabular}{l l c c c c}
\toprule
\textbf{Model} &
\textbf{Backbone} &
\textbf{mIoU $\uparrow$} &
\textbf{mAcc $\uparrow$} &
\textbf{mDice $\uparrow$} &
\textbf{aAcc $\uparrow$} \\
\midrule

FCN &
U-Net \cite{ronneberger2015u} &
73.17 $\pm$ 1.27 &
86.67 $\pm$ 4.00 &
82.13 $\pm$ 1.06 &
97.77 $\pm$ 0.45 \\

DeepLabV3 \cite{chen2017rethinking} &
ResNet50 \cite{he2016deep} &
75.49 $\pm$ 0.46 &
91.12 $\pm$ 0.12 &
84.15 $\pm$ 0.37 &
97.92 $\pm$ 0.08 \\

DeepLabV3+ \cite{deeplabv3plus2018} &
ResNet50 \cite{he2016deep} &
\underline{76.25 $\pm$ 0.24} &
90.65 $\pm$ 0.59 &
\underline{84.77 $\pm$ 0.20} &
\underline{98.06 $\pm$ 0.01} \\

SegFormer \cite{xie2021segformer} &
MiT-B0 &
73.95 $\pm$ 1.03 &
\underline{92.35 $\pm$ 1.48} &
82.87 $\pm$ 0.86 &
97.59 $\pm$ 0.27 \\

SegFormer \cite{xie2021segformer} &
MiT-B2 &
75.74 $\pm$ 0.51 &
90.96 $\pm$ 1.03 &
84.35 $\pm$ 0.41 &
97.96 $\pm$ 0.13 \\

UPerNet \cite{xiao2018unified} &
ResNet50 \cite{he2016deep} &
\textbf{76.49 $\pm$ 0.40} &
90.33 $\pm$ 0.04 &
\textbf{84.96 $\pm$ 0.32} &
\textbf{98.11 $\pm$ 0.06} \\

UPerNet \cite{xiao2018unified} &
ConvNeXt-Tiny \cite{liu2022convnet} &
74.95 $\pm$ 0.76 &
\textbf{93.08 $\pm$ 0.33} &
83.72 $\pm$ 0.64 &
97.74 $\pm$ 0.15 \\

\bottomrule
\end{tabular}
\color{black}
\end{table}

As shown in Table~\ref{tab:gingiva_benchmark}, UPerNet with ResNet50 achieved the highest mIoU
(76.49\%), mDice (84.96\%), and overall pixel accuracy (98.11\%) among the evaluated
configurations. DeepLabV3+ with ResNet50 produced similar results, with an mIoU of 76.25\% and
an mDice of 84.77\%. UPerNet with ConvNeXt-Tiny achieved the highest mAcc (93.08\%), followed
by SegFormer with MiT-B0 (92.35\%). The ranking of the models therefore varied according to the
evaluation metric. No configuration achieved the highest value for every metric. In particular,
metrics measuring region overlap and performance for the gingivitis class provide information
that cannot be captured by overall pixel accuracy alone.


\begin{figure}[!htbp]
\centering
\includegraphics[width=\textwidth]
{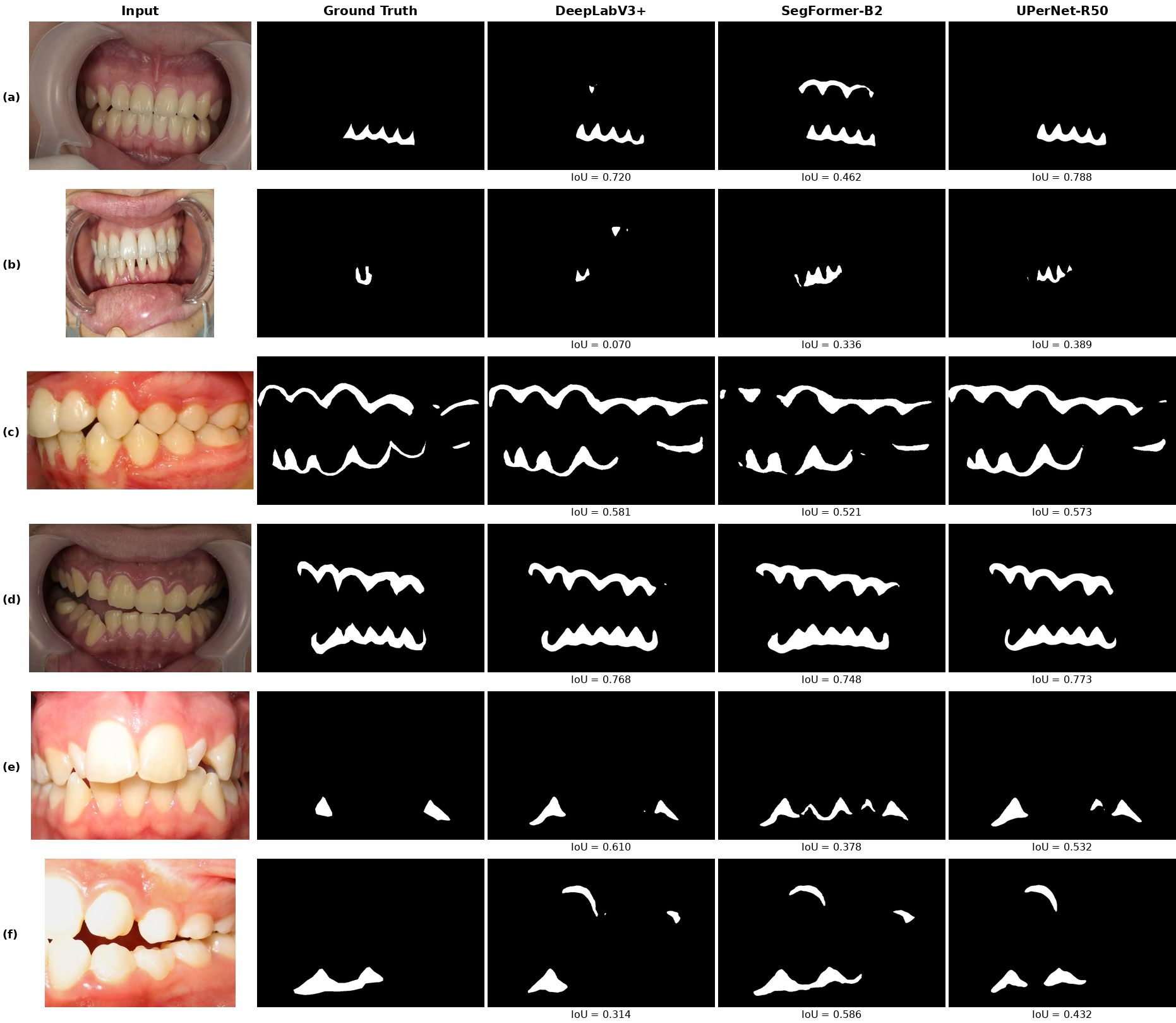}

\caption{
Qualitative comparison of DeepLabV3+ with ResNet50,
SegFormer with MiT-B2, and UPerNet with ResNet50
on selected test images. Each row presents the
input image, reference gingivitis mask, and
predictions obtained from the three configurations.
Gingivitis IoU is reported below each prediction.
The selected images illustrate differences in
the size of gingivitis regions, boundary complexity,
and segmentation results between models.
}
\label{fig:qualitative_comparison}
\end{figure}

\FloatBarrier

Figure~\ref{fig:qualitative_comparison} provides a visual comparison of DeepLabV3+ with
ResNet50, SegFormer with MiT-B2, and UPerNet with ResNet50 on selected test images. The
examples represent different sizes of gingivitis regions and different boundary
characteristics. All three configurations identified the principal gingivitis regions in the
selected examples, although differences were more apparent in some images containing smaller
areas of inflammation. The comparison also shows that segmentation performance varies between
individual images and that different configurations may produce more accurate predictions in
different cases.


\begin{table}[htbp]
\centering
\color{black}
\caption{
Segmentation performance for the background
and gingivitis classes on the GingivitisSeg dataset.
Results are presented as mean $\pm$ standard deviation.
The subscripts ``back'' and ``gin'' denote the
background and gingivitis classes, respectively.
Bold values indicate the highest performance,
and underlined values indicate the second highest.
All metrics are reported as percentages.
}
\label{tab:classwise_results}

\resizebox{\textwidth}{!}{
\begin{tabular}{l l c c c c c c}
\toprule
\textbf{Model} &
\textbf{Backbone} &
\multicolumn{3}{c}{\textbf{Background}} &
\multicolumn{3}{c}{\textbf{Gingivitis}} \\
\cmidrule(lr){3-5}
\cmidrule(lr){6-8}
& &
\textbf{IoU$_{back}$ $\uparrow$} &
\textbf{Acc$_{back}$ $\uparrow$} &
\textbf{Dice$_{back}$ $\uparrow$} &
\textbf{IoU$_{gin}$ $\uparrow$} &
\textbf{Acc$_{gin}$ $\uparrow$} &
\textbf{Dice$_{gin}$ $\uparrow$} \\
\midrule

FCN &
U-Net \cite{ronneberger2015u} &
97.72 $\pm$ 0.46 &
98.42 $\pm$ 0.71 &
98.85 $\pm$ 0.23 &
48.63 $\pm$ 2.09 &
74.92 $\pm$ 8.70 &
65.42 $\pm$ 1.88 \\

DeepLabV3 \cite{chen2017rethinking} &
ResNet50 \cite{he2016deep} &
97.87 $\pm$ 0.08 &
98.33 $\pm$ 0.09 &
98.92 $\pm$ 0.04 &
53.11 $\pm$ 0.84 &
83.90 $\pm$ 0.33 &
69.37 $\pm$ 0.71 \\

DeepLabV3+ \cite{deeplabv3plus2018} &
ResNet50 \cite{he2016deep} &
\underline{98.02 $\pm$ 0.01} &
\underline{98.50 $\pm$ 0.03} &
\underline{99.00 $\pm$ 0.01} &
\underline{54.50 $\pm$ 0.47} &
82.78 $\pm$ 1.20 &
\underline{70.55 $\pm$ 0.40} \\

SegFormer \cite{xie2021segformer} &
MiT-B0 &
97.53 $\pm$ 0.27 &
97.90 $\pm$ 0.37 &
98.75 $\pm$ 0.14 &
50.37 $\pm$ 1.79 &
\underline{86.80 $\pm$ 3.32} &
66.99 $\pm$ 1.58 \\

SegFormer \cite{xie2021segformer} &
MiT-B2 &
97.92 $\pm$ 0.13 &
98.38 $\pm$ 0.20 &
98.95 $\pm$ 0.06 &
53.57 $\pm$ 0.88 &
83.54 $\pm$ 2.25 &
69.76 $\pm$ 0.75 \\

UPerNet \cite{xiao2018unified} &
ResNet50 \cite{he2016deep} &
\textbf{98.06 $\pm$ 0.06} &
\textbf{98.57 $\pm$ 0.06} &
\textbf{99.02 $\pm$ 0.03} &
\textbf{54.91 $\pm$ 0.74} &
82.09 $\pm$ 0.03 &
\textbf{70.89 $\pm$ 0.62} \\

UPerNet \cite{xiao2018unified} &
ConvNeXt-Tiny \cite{liu2022convnet} &
97.68 $\pm$ 0.15 &
98.01 $\pm$ 0.17 &
98.83 $\pm$ 0.08 &
52.24 $\pm$ 1.38 &
\textbf{88.14 $\pm$ 0.85} &
68.62 $\pm$ 1.19 \\

\bottomrule
\end{tabular}
}
\color{black}
\end{table}

Table~\ref{tab:classwise_results} presents performance separately for the background and
gingivitis classes. All evaluated configurations achieved higher IoU and Dice scores for the
background than for gingivitis. UPerNet with ResNet50 obtained the highest gingivitis IoU
(54.91\%) and Dice (70.89\%). UPerNet with ConvNeXt-Tiny achieved the highest accuracy for the
gingivitis class (88.14\%). DeepLabV3+ with ResNet50 also achieved similar segmentation
performance for gingivitis, with an IoU of 54.50\% and Dice of 70.55\%. These findings
demonstrate the importance of reporting performance for the gingivitis class rather than
relying only on overall pixel accuracy, which is strongly influenced by the background.


\begin{table}[htbp]
\centering
\color{black}
\caption{
Statistical comparisons between UPerNet with ResNet50
and the remaining segmentation configurations.
Values are $p$-values obtained from paired Wilcoxon
signed rank tests on scores calculated separately
for each of the 145 test images, with adjustment
for multiple comparisons using the Holm procedure.
Bold values indicate statistically significant
differences ($p_{\mathrm{Holm}}<0.05$).
}
\label{tab:statistical_comparison}
\footnotesize
\renewcommand{\arraystretch}{1.15}

\begin{tabular}{lcccc}
\toprule
\textbf{Comparator} &
\textbf{mIoU} &
\textbf{mDice} &
\textbf{IoU$_{\mathrm{gin}}$} &
\textbf{Dice$_{\mathrm{gin}}$} \\
\midrule

FCN/U-Net &
\textbf{$<0.001$} &
\textbf{$<0.001$} &
\textbf{$<0.001$} &
\textbf{$<0.001$} \\

DeepLabV3 with ResNet50 &
\textbf{0.004} &
\textbf{0.007} &
\textbf{0.005} &
\textbf{0.009} \\

DeepLabV3+ with ResNet50 &
0.730 & 0.797 & 0.637 & 0.741 \\

SegFormer with MiT-B0 &
\textbf{$<0.001$} &
\textbf{$<0.001$} &
\textbf{$<0.001$} &
\textbf{$<0.001$} \\

SegFormer with MiT-B2 &
0.0518 & 0.0503 & 0.0576 & 0.0567 \\

UPerNet with ConvNeXt-Tiny &
\textbf{0.009} &
\textbf{0.024} &
\textbf{0.031} &
\textbf{0.042} \\

\bottomrule
\end{tabular}
\color{black}
\end{table}

The paired Wilcoxon signed rank tests identified statistically significant differences for
several model comparisons. As shown in Table~\ref{tab:statistical_comparison}, UPerNet with
ResNet50 differed significantly from FCN/U-Net, DeepLabV3 with ResNet50, SegFormer with MiT-B0,
and UPerNet with ConvNeXt-Tiny for all four evaluated overlap metrics after Holm correction.

For gingivitis segmentation, the mean difference in IoU scores calculated for individual images
between UPerNet with ResNet50 and FCN/U-Net was 7.73 percentage points. The corresponding
difference in Dice was 8.06 percentage points. Compared with DeepLabV3 with ResNet50, the mean
differences were 2.01 points for IoU and 1.97 points for Dice. The corresponding differences
were 3.36 and 2.97 points for SegFormer with MiT-B0, and 1.21 and 0.91 points for UPerNet with
ConvNeXt-Tiny.

In contrast, the differences between UPerNet with ResNet50 and DeepLabV3+ with ResNet50 were
not statistically significant for any of the four evaluated metrics. The differences relative
to SegFormer with MiT-B2 were also not significant after Holm correction. Therefore, although
UPerNet with ResNet50 achieved the highest numerical mIoU and mDice, its advantage over the
closest competing configurations should not be interpreted as statistically established
superiority.


\begin{table}[!htbp]
\centering
\caption{
Summary of representative studies using artificial
intelligence for gingivitis analysis.
The studies differ in datasets, tasks, annotations,
and evaluation procedures. Their reported results
are presented for descriptive purposes and should
not be interpreted as direct performance comparisons.
}
\label{tab:reported_iou_accuracy_comparison}

\resizebox{\textwidth}{!}{
\small
\renewcommand{\arraystretch}{1.15}

\begin{tabular}{p{3.2cm} p{2.6cm} p{3.4cm} c c}
\toprule
\textbf{Study} &
\textbf{Task} &
\textbf{Model} &
\textbf{IoU / mIoU} &
\textbf{Accuracy} \\
\midrule

Li et al. (2021) \cite{li2021automatic} &
Gingival disease segmentation &
DeepLabV3+ &
0.650$^{a}$ / 0.379--0.485$^{b}$ &
-- \\

Chau et al. (2023) \cite{chau2023accuracy} &
Gingivitis detection &
Deep learning segmentation model &
0.600 &
-- \\

Cheng et al. (2024) \cite{cheng2024developing} &
Mobile gingivitis detection &
Mobile deep learning model &
-- &
$>$0.900 \\

Chau et al. (2025) \cite{chau2025external} &
External validation &
Deep learning gingivitis detection model &
-- &
0.850 \\

Cheng et al. (2025) \cite{cheng2025gumagent} &
Gingival disease segmentation &
GumAgent &
0.482 &
0.886$^{c}$ \\

Liu et al. (2024) \cite{liu2024oral} &
Segmentation of multiple oral diseases &
Oral-Mamba &
0.700$^{d}$ &
0.830$^{d}$ \\

\midrule

\textbf{Ours} &
\textbf{Gingivitis segmentation} &
\textbf{UPerNet with ResNet50} &
\textbf{0.549$^{e}$ / 0.765$^{f}$} &
\textbf{0.981$^{g}$} \\

\bottomrule
\end{tabular}
}

\vspace{1mm}

\footnotesize{
$^{a}$ Reported result for binary segmentation of diseased gingiva.\\
$^{b}$ Reported mIoU range for gingival condition segmentation
with four and five classes.\\
$^{c}$ Reported image validation accuracy for the GumAgent workflow.\\
$^{d}$ Reported result for the gingivitis class using Oral-Mamba.\\
$^{e}$ IoU for the gingivitis class in this study.\\
$^{f}$ Mean IoU across background and gingivitis classes
in this study.\\
$^{g}$ Overall pixel accuracy in this study.
}
\end{table}

Table~\ref{tab:reported_iou_accuracy_comparison} provides an overview of representative studies
on gingivitis analysis. The reported performance values cannot be compared directly because the
studies use different image sources, segmentation targets, annotation procedures, numbers of
classes, and evaluation protocols. In the GingivitisSeg benchmark, UPerNet with ResNet50
achieved an mIoU of 0.765, an IoU of 0.549 for the gingivitis class, and an overall pixel
accuracy of 0.981. These results describe performance within the present benchmark and do not
establish superiority over studies using different datasets.

Overall, the benchmark results varied with the evaluated segmentation configurations. UPerNet
with ResNet50 achieved the highest numerical values for aggregate overlap metrics, whereas
other configurations achieved higher values for individual metrics. For example, UPerNet with
ConvNeXt-Tiny obtained the highest accuracy for the gingivitis class. The statistical
comparisons also showed that some numerical differences between the strongest configurations
were not significant. These findings support a cautious interpretation of small differences in
model rankings and demonstrate the importance of evaluating both aggregate performance and
segmentation quality for the gingivitis class.

\subsection{
Ablation Study of Optimization Components}
\label{sec:ablation}

An ablation study was performed using UPerNet with a ResNet50 backbone to evaluate the
contribution of individual optimization components. The complete configuration combined Cross
Entropy loss with class weighting, Dice loss, and auxiliary supervision. Each component was
removed separately while the remaining experimental settings were kept unchanged. A
configuration using only standard Cross Entropy loss was also included as a baseline. For every
configuration, the checkpoint with the highest validation mIoU was selected for evaluation on
the test set.

\begin{table}[htbp]
\centering
\color{black}
\caption{
Ablation study of optimization components using
UPerNet with a ResNet50 backbone.
Class Wt. denotes class weighting, and Aux. Sup.
denotes auxiliary supervision. Check marks indicate
the components included in each configuration.
All metrics are reported as percentages.
}
\label{tab:ablation}
\footnotesize
\renewcommand{\arraystretch}{1.18}
\setlength{\tabcolsep}{6pt}

\begin{tabular}{l c c c c c c c}
\toprule
&
\multicolumn{3}{c}{
\shortstack{\textbf{Optimization}\\\textbf{Components}}
}
&
\multicolumn{2}{c}{
\shortstack{\textbf{Overall}\\\textbf{Performance}}
}
&
\multicolumn{2}{c}{
\shortstack{\textbf{Gingivitis}\\\textbf{Class}}
} \\
\cmidrule(lr){2-4}
\cmidrule(lr){5-6}
\cmidrule(lr){7-8}

\textbf{Configuration} &
\shortstack{\textbf{Class}\\\textbf{Wt.}} &
\shortstack{\textbf{Dice}\\\textbf{Loss}} &
\shortstack{\textbf{Aux.}\\\textbf{Sup.}} &
\textbf{mIoU} &
\textbf{mDice} &
\textbf{IoU$_{gin}$} &
\textbf{Dice$_{gin}$} \\
\midrule

\textbf{Full} &
$\checkmark$ & $\checkmark$ & $\checkmark$ &
\textbf{76.77} & \textbf{85.18} &
\textbf{55.43} & \textbf{71.33} \\

w/o Class Wt. &
$\times$ & $\checkmark$ & $\checkmark$ &
75.90 & 84.41 & 53.45 & 69.66 \\

w/o Dice Loss &
$\checkmark$ & $\times$ & $\checkmark$ &
73.63 & 82.61 & 49.87 & 66.55 \\

w/o Aux. Sup. &
$\checkmark$ & $\checkmark$ & $\times$ &
75.31 & 84.01 & 52.83 & 69.14 \\

CE Only &
$\times$ & $\times$ & $\times$ &
75.15 & 83.79 & 52.05 & 68.46 \\

\bottomrule
\end{tabular}
\color{black}
\end{table}

As shown in Table~\ref{tab:ablation}, the complete configuration achieved the highest
performance for all reported overlap metrics, with an mIoU of 76.77\%, mDice of 85.18\%,
gingivitis IoU of 55.43\%, and gingivitis Dice of 71.33\%. Removing Dice loss produced the
largest decrease in performance. Gingivitis IoU decreased from 55.43\% to 49.87\%, while Dice
decreased from 71.33\% to 66.55\%. Removing auxiliary supervision also reduced performance for
the gingivitis class, yielding an IoU of 52.83\% and Dice of 69.14\%. Removing class weighting
resulted in a smaller decrease, with gingivitis IoU and Dice values of 53.45\% and 69.66\%,
respectively.

The configuration using only Cross Entropy loss achieved an mIoU of 75.15\%, gingivitis IoU of
52.05\%, and gingivitis Dice of 68.46\%. The results indicate that the combination of the
evaluated optimization components achieved the highest segmentation performance in this
experiment. Among the individual components, removing Dice loss produced the largest
performance reduction.

\subsection{Analysis of Foreground Size}
\label{sec:foreground_size}

To examine the relationship between the size of gingivitis regions and segmentation
performance, an additional analysis was performed using UPerNet with a ResNet50 backbone. For
each test image, the foreground area was calculated as the percentage of valid pixels belonging
to the gingivitis class in the reference mask. The corresponding gingivitis IoU and Dice scores
were then calculated separately for each image. All 145 test images contained gingivitis
regions and were included in the analysis. The foreground proportion ranged from 0.14\% to
19.83\%, with a mean of 3.55\%.

\begin{figure}[htbp]
\centering
\includegraphics[width=0.90\linewidth]
{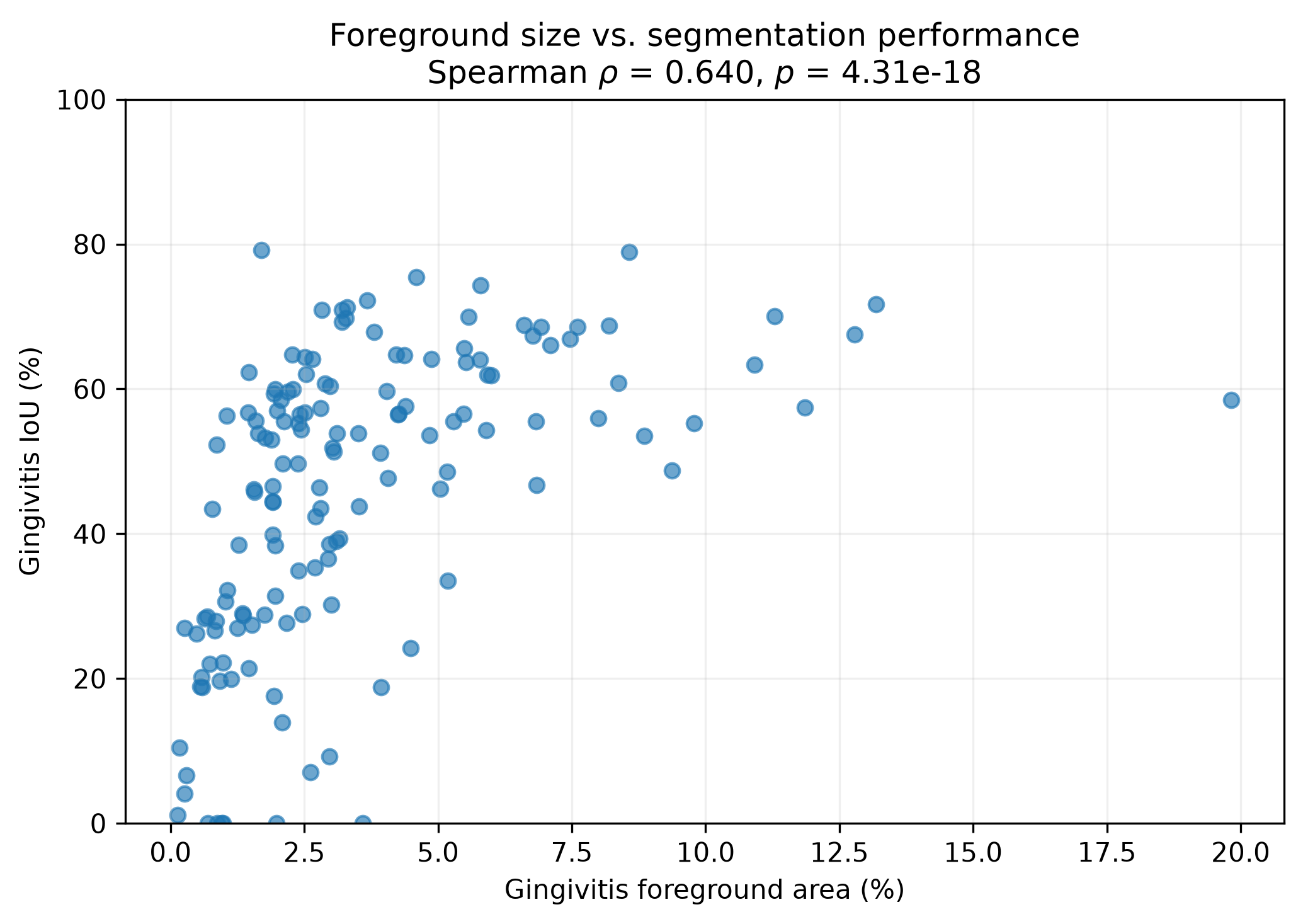}
\caption{
Relationship between gingivitis region size and
segmentation performance on the test set.
Foreground size is defined as the percentage
of pixels belonging to gingivitis in the
reference mask. A positive association was
observed between region size and gingivitis IoU
(Spearman's $\rho=0.640$,
$p=4.31\times10^{-18}$).
}
\label{fig:foreground_size_iou}
\end{figure}

As shown in Fig.~\ref{fig:foreground_size_iou}, segmentation performance was positively
associated with the relative area occupied by gingivitis. Spearman's correlation analysis
showed a statistically significant positive association between foreground area and gingivitis
IoU ($\rho=0.640$, $p=4.31\times10^{-18}$). Images containing smaller gingivitis regions
generally exhibited lower IoU scores and greater variability. Larger regions were associated
with higher IoU values. To examine this relationship further, the 145 test images were divided
into three groups according to the proportion of pixels belonging to gingivitis. The groups
represented small, medium, and large foreground regions and were defined using tertiles of the
foreground proportions. Table~\ref{tab:foreground_size_groups} summarizes the results.

\begin{table}[htbp]
\centering
\color{black}
\caption{
Segmentation performance according to gingivitis
region size. Images were divided into three groups
using tertiles of foreground area.
Values are reported as mean $\pm$ standard deviation.
}
\label{tab:foreground_size_groups}
\footnotesize
\renewcommand{\arraystretch}{1.15}

\begin{tabular}{lcccc}
\toprule
\textbf{Foreground Group} &
\textbf{$n$} &
\textbf{Mean Area (\%)} &
\textbf{IoU$_{gin}$ (\%)} &
\textbf{Dice$_{gin}$ (\%)} \\
\midrule

Small &
48 & 1.13 &
31.30 $\pm$ 19.01 &
44.47 $\pm$ 22.92 \\

Medium &
48 & 2.66 &
47.00 $\pm$ 18.50 &
61.43 $\pm$ 20.67 \\

Large &
49 & 6.79 &
59.38 $\pm$ 11.85 &
73.74 $\pm$ 10.74 \\

\bottomrule
\end{tabular}
\color{black}
\end{table}

Segmentation performance increased across the groups defined by foreground size. Mean
gingivitis IoU increased from 31.30\% for small regions to 47.00\% for medium regions and
59.38\% for large regions. Similarly, mean gingivitis Dice increased from 44.47\% to 61.43\%
and 73.74\%, respectively. Performance also showed greater variability among images with
smaller gingivitis regions, as indicated by the standard deviations of IoU and Dice.

These findings indicate an association between gingivitis region size and segmentation
performance. Small regions were generally more difficult to delineate than larger regions. This
observation is consistent with the pronounced imbalance between gingivitis and background
pixels in the dataset. The results highlight the importance of evaluating segmentation
performance on small gingivitis regions rather than relying predominantly on aggregate pixel
accuracy.

\subsection{
Sensitivity to Aspect Ratio Preservation}
\label{sec:aspect_ratio}

To investigate the influence of image resizing, a sensitivity experiment was conducted using
UPerNet with ResNet50. The reference configuration directly resized images to $512\times512$
pixels without preserving the original aspect ratio. In the alternative configuration, the
original aspect ratio was preserved during resizing. Padding was then applied to obtain the
same $512\times512$ network input dimensions. All other model settings, dataset partitions,
augmentation procedures, and evaluation criteria were kept unchanged.

\begin{table}[htbp]
\centering
\color{black}
\caption{
Sensitivity analysis of image resizing using UPerNet
with ResNet50. The configurations differ in whether
the original image aspect ratio is preserved.
All metrics are reported as percentages.
}
\label{tab:aspect_ratio}
\footnotesize
\renewcommand{\arraystretch}{1.15}

\begin{tabular}{lccccccc}
\toprule
\textbf{Preprocessing} &
\textbf{mIoU} &
\textbf{mAcc} &
\textbf{mDice} &
\textbf{aAcc} &
\textbf{IoU$_{\mathrm{gin}}$} &
\textbf{Acc$_{\mathrm{gin}}$} &
\textbf{Dice$_{\mathrm{gin}}$} \\
\midrule

Direct resize ($512\times512$) &
\textbf{76.77} &
\textbf{90.36} &
\textbf{85.18} &
\textbf{98.15} &
\textbf{55.43} &
\textbf{82.11} &
\textbf{71.33} \\

Aspect ratio preservation with padding &
74.41 & 89.84 & 83.23 & 97.81 &
51.06 & 81.41 & 67.60 \\

$\Delta$ &
$-2.36$ & $-0.52$ & $-1.95$ & $-0.34$ &
$-4.37$ & $-0.70$ & $-3.73$ \\

\bottomrule
\end{tabular}
\color{black}
\end{table}

As shown in Table~\ref{tab:aspect_ratio}, preserving the original image aspect ratio did not
improve segmentation performance in the evaluated configuration. Compared with direct resizing,
the alternative method reduced mIoU from 76.77\% to 74.41\% and mDice from 85.18\% to 83.23\%.
The reduction was more pronounced for gingivitis, with IoU decreasing from 55.43\% to 51.06\%
and Dice decreasing from 71.33\% to 67.60\%.

Direct resizing therefore produced better results than resizing with aspect ratio preservation
and padding in this experiment. However, this finding does not demonstrate that geometric
distortion is generally preferable. The result is specific to the evaluated dataset, model
configuration, and preprocessing procedures.

\subsection{Robustness to Image Blur}
\label{sec:blur_robustness}

Additional experiments were conducted to evaluate the sensitivity of gingivitis segmentation to
degradation in image quality. UPerNet with ResNet50 was evaluated without retraining on all 145
test images after applying synthetic Gaussian and motion blur. Gaussian blur was applied at
three levels, with $\sigma=1$, $\sigma=2$, and $\sigma=3$. Motion blur was generated using
directional kernels of 5, 9, and 15 pixels at an angle of $45^{\circ}$. The original test
images were used as the reference condition without degradation.

For each condition, gingivitis IoU and Dice scores were calculated separately for individual
test images. Because the same images were evaluated under the original and degraded conditions,
paired comparisons were performed using Wilcoxon signed rank tests. The resulting p-values were
adjusted using the Holm procedure to account for multiple comparisons across degradation
levels. The results are presented in Table~\ref{tab:blur_robustness}.

\begin{table}[htbp]
\centering
\color{black}
\caption{
Segmentation performance under Gaussian and motion blur.
Changes ($\Delta$) are reported relative to the original
images. Statistical comparisons were performed using
paired Wilcoxon signed rank tests with Holm correction.
All performance values are percentages.
}
\label{tab:blur_robustness}
\footnotesize
\renewcommand{\arraystretch}{1.12}

\begin{tabular}{lccc|ccc}
\toprule
&
\multicolumn{3}{c|}{\textbf{Gingivitis IoU}} &
\multicolumn{3}{c}{\textbf{Gingivitis Dice}} \\
\cmidrule(lr){2-4}
\cmidrule(lr){5-7}

\textbf{Condition} &
\textbf{Mean} &
\textbf{$\Delta$} &
\textbf{$p_{\mathrm{Holm}}$} &
\textbf{Mean} &
\textbf{$\Delta$} &
\textbf{$p_{\mathrm{Holm}}$} \\
\midrule

Clean &
45.99 & -- & -- &
59.98 & -- & -- \\

Gaussian ($\sigma=1$) &
45.09 & $-0.90$ & 0.646 &
58.81 & $-1.16$ & 0.699 \\

Gaussian ($\sigma=2$) &
41.94 & $-4.04$ & \textbf{0.025} &
54.94 & $-5.04$ & \textbf{0.030} \\

Gaussian ($\sigma=3$) &
38.60 & $-7.39$ & \textbf{0.0011} &
50.74 & $-9.24$ & \textbf{0.00085} \\

Motion ($k=5$) &
44.25 & $-1.74$ & 0.646 &
57.55 & $-2.43$ & 0.699 \\

Motion ($k=9$) &
40.26 & $-5.72$ & \textbf{0.012} &
52.73 & $-7.25$ & \textbf{0.0107} \\

Motion ($k=15$) &
35.08 & $-10.91$ &
\textbf{$3.48\times10^{-6}$} &
46.15 & $-13.82$ &
\textbf{$1.78\times10^{-6}$} \\

\bottomrule
\end{tabular}
\color{black}
\end{table}

Mild image blur produced relatively small changes in segmentation performance. Under Gaussian
blur with $\sigma=1$, the mean gingivitis IoU decreased by 0.90 percentage points, while Dice
decreased by 1.16 points. Neither difference was statistically significant after Holm
correction. Similarly, motion blur with a kernel size of 5 pixels reduced IoU by 1.74 points
and Dice by 2.43 points, without statistically significant differences from the original
images.

Stronger blur produced progressively larger reductions in performance. Under Gaussian blur with
$\sigma=2$, mean gingivitis IoU decreased from 45.99\% to 41.94\%, while Dice decreased from
59.98\% to 54.94\%. Both reductions were statistically significant after Holm correction. At
$\sigma=3$, the reductions reached 7.39 percentage points for IoU and 9.24 points for Dice.

Motion blur produced a similar pattern. A kernel size of 9 pixels reduced IoU by 5.72 points
and Dice by 7.25 points. The largest reduction occurred with a motion blur kernel of 15 pixels.
Under this condition, mean IoU decreased to 35.08\% and mean Dice to 46.15\%. These reductions
were statistically significant after Holm correction ($p_{\mathrm{Holm}}<0.001$).

Overall, the evaluated model showed relatively small performance changes under mild blur but
became increasingly sensitive as the degree of degradation increased, particularly under strong
motion blur. These observations indicate that image quality can influence gingivitis
segmentation performance. However, the experiments used synthetic image degradation. The
findings cannot establish robustness under all conditions encountered in clinical photography
or smartphone imaging.

\subsection{Limitations}
\label{sec:limitations}

Several limitations of this study should be acknowledged. First, the reference segmentation
masks were created according to visible signs of gingival inflammation, including redness,
edema, and changes in gingival contour. Clinical periodontal indices such as the Gingival Index
(GI), Modified Gingival Index (MGI), and bleeding on probing (BOP) were not used to validate
the segmentation masks. Therefore, the benchmark evaluates segmentation of visually identified
gingival inflammation rather than clinically validated disease severity.

Second, demographic information and detailed smartphone device characteristics were not
recorded for the 432 photographs collected at Ankara University. Consequently, demographic
subgroup analyses and comparisons between acquisition devices could not be performed. Although
the annotation process included expert calibration and joint consensus review, the same images
were not independently annotated by both experts. Therefore, quantitative agreement between the
annotators could not be calculated. Consistency within the same annotator was assessed
separately using 30 photographs annotated again after one month, yielding an ICC of 0.80.

Third, reliable patient identifiers were unavailable for the additional 529 photographs
obtained from the public dataset. Consequently, complete independence between patients across
the training, validation, and test partitions could not be established. Furthermore, all images
included annotated gingivitis regions. The benchmark therefore does not evaluate the ability to
distinguish healthy photographs from images containing gingivitis.

Finally, the robustness experiments were based on synthetic Gaussian and motion blur rather
than naturally degraded clinical photographs. The observed results should therefore not be
interpreted as evidence of performance under all conditions encountered in clinical imaging.
Independent clinical validation remains necessary before considering the deployment of these
segmentation models in clinical practice.

\section{Conclusions}\label{sec:conclusions}

This study presented the expanded GingivitisSeg benchmark for automatic gingivitis segmentation
from intraoral photographs. The expanded dataset comprises 961 intraoral images from two image
sources, with expert annotated pixel level gingivitis masks and predefined training,
validation, and test splits. By focusing specifically on gingivitis segmentation, the dataset
addresses an important gap in the literature, where most previous studies have focused on
detection, localization, or clinical diagnosis rather than detailed pixel-level segmentation.

A comprehensive benchmark evaluation was conducted using representative CNN-based,
transformer-based, and multi scale segmentation architectures. The evaluated configurations
used the same dataset partitions, input size, and evaluation criteria, while architecture
specific training settings were retained where required. UPerNet--ResNet50 achieved the highest
mIoU (76.49\%) and mDice (84.96\%), while controlled ablation experiments supported the
contribution of the foreground-aware optimization components.

Despite the promising results, gingivitis segmentation remains challenging because of small
foreground regions, low-contrast boundaries, illumination variation, and image quality
degradation. Future work may focus on larger and externally validated datasets, patient level
evaluation, boundary aware learning, and improved segmentation of small gingivitis regions.
Future studies may also investigate the relationship between clinically assessed gingivitis
severity and segmentation performance, including whether different inflammation grades
influence the accuracy of gingival region delineation. Overall, GingivitisSeg provides a useful
public benchmark for advancing deep learning-based gingivitis segmentation and computer-aided
periodontal assessment.

\section*{Acknowledgements}
Not applicable.

\section*{Availability of data and materials}

The GingivitisSeg dataset generated and analyzed during this study is publicly available in the Zenodo repository at \url{https://doi.org/10.5281/zenodo.21098897}. The repository includes the intraoral images, expert-annotated pixel-level segmentation masks, predefined training, validation, and test splits, and accompanying documentation.

\section*{Code availability}

The source code, benchmark implementation, and documentation are publicly available at:
\url{https://github.com/zeynepozdemir/GingivitisSeg}

\section*{Funding}

The authors declare that no funds, grants, or other support were received during the preparation of this study.

\section*{Declarations}

\subsection*{Ethics approval and consent to participate}

The original 432-image clinical subset was approved by the Clinical
Research Ethics Committee of Ankara University Faculty of Dentistry
(Decision No. 05/02, February 27, 2023), and written informed consent was
obtained prior to image acquisition. The additional 529 images were obtained
from a publicly available dataset and used as secondary data
\cite{duy2024dental}.

\subsection*{Consent for publication}

Written informed consent for publication was obtained for the original
clinical subset. The additional images were obtained from a publicly
available source.

\subsection*{Competing interests}

The authors have no relevant financial or non-financial interests to disclose.

\section*{Author contributions}
Zeynep Ozdemir conceived and designed the study, developed the methodology, curated the
 dataset for deep learning experiments, established the benchmark protocol, conducted the
 experiments, analyzed the results, and wrote the original draft of the manuscript.
Irem Ulku and Omer Ozgur Tanr\i over contributed to the methodology, provided technical
feedback, and critically reviewed and revised the manuscript.
Zehra Hasgul and Fatma Karacao\u{g}lu performed the clinical annotation of the gingivitis
images, verified the ground-truth masks, and reviewed the manuscript with respect to its
clinical content. Murat Akkaya contributed to the acquisition and development of the clinical
dataset. All authors read and approved the final manuscript.

\bibliographystyle{unsrt}
\bibliography{references}

\end{document}